\documentclass{article} 

\usepackage[preprint,nonatbib]{neurips_2020}
\usepackage[utf8]{inputenc} % allow utf-8 input
\usepackage[T1]{fontenc}    % use 8-bit T1 fonts
\usepackage{hyperref}       % hyperlinks
\usepackage{url}            % simple URL typesetting
\usepackage{booktabs}       % professional-quality tables
\usepackage{amsfonts}       % blackboard math symbols
\usepackage{nicefrac}       % compact symbols for 1/2, etc.
\usepackage{microtype}      % microtypography

\usepackage{color}
\usepackage{bm}
\usepackage{etoolbox}
\usepackage{enumerate}
\usepackage{amsmath}
\usepackage{layouts}
\usepackage{graphicx}
\usepackage{algorithm}
\usepackage[noend]{algpseudocode}
\usepackage{subcaption}
\usepackage{multirow}

\usepackage{comment}
\usepackage{xcolor}
\usepackage{etoolbox}

\newcommand{\df}{Deepfake}

\newbool{inccomment}
\setbool{inccomment}{true}
\newcommand{\cy}[1]{\ifbool{inccomment}{{\color{red} #1}}{}}

\title{Defending against GAN-based Deepfake Attacks via Transformation-aware Adversarial Faces}

\author{
  Chaofei Yang\textsuperscript{1}, Lei Ding\textsuperscript{2}, Yiran Chen\textsuperscript{1}, Hai Li\textsuperscript{1} \\
  \textsuperscript{1}Duke University, \textsuperscript{2}Accenture Labs \\
  \textsuperscript{1}\texttt{\{chaofei.yang, yiran.chen, hai.li\}@duke.edu} \\
  \textsuperscript{2}\texttt{lei.a.ding@accenture.com} \\
}

\begin{document}

\maketitle

\begin{abstract}
\df{} represents a category of face-swapping attacks that leverage machine learning models such as autoencoders or generative adversarial networks.
Although the concept of the face-swapping is not new, its recent technical advances make fake content (e.g., images, videos) more realistic and imperceptible to Humans.
Various detection techniques for \df{} attacks have been explored.
These methods, however, are passive measures against \df{}s as they are mitigation strategies after the high-quality fake content is generated.
More importantly, we would like to think ahead of the attackers with robust defenses. 
This work aims to take an offensive measure to impede the generation of high-quality fake images or videos. 
Specifically, we propose to use novel transformation-aware adversarially perturbed faces as a defense against GAN-based \df{} attacks. 
Different from the na\"ive adversarial faces, our proposed approach leverages differentiable random image transformations during the generation.
We also propose to use an ensemble-based approach to enhance the defense robustness against GAN-based \df{} variants under the black-box setting.
We show that training a \df{} model with adversarial faces can lead to a significant degradation in the quality of synthesized faces. 
This degradation is twofold.
On the one hand, the quality of the synthesized faces is reduced with more visual artifacts such that the synthesized faces are more obviously fake or less convincing to human observers.
On the other hand, the synthesized faces can easily be detected based on various metrics.
\end{abstract}
\section{Introduction}
\label{sec:intro}

Machine learning (ML) has experienced rapid development during the past decade and been widely adopted by many daily applications.
The adoption of ML technologies, however, also induces new threats in data privacy and security. 
As a famous example, \df{}~\cite{deepfake_original} recently draws increasing attention by offering the capability to generate a fake video of any particular person:
in the fake video, an attacker can swap a person's face with the synthesized face of another person who can be anyone such as a celebrity, a politician, or just a normal person.
Although the concept of face-swap has been studied for long, 
the threat of \df{} becomes much severer because of the utilization of generative models such as autoencoders~\cite{kramer1991nonlinear} and generative adversarial networks (GANs)~\cite{goodfellow2014generative}. 
These techniques enable the generation of extremely realistic synthetic images with incredible details.
The faces are well synthesized such that naked eyes cannot easily distinguish between a fake video and an authentic one.
\df{} can be potentially used for deceiving identity verification or defaming a person.

Many algorithms have been proposed to detect \df{}  videos~\cite{guera2018deepfake,li2018exposing,rossler2019faceforensics++}.
These algorithms usually adopt state-of-the-art neural network (NN) models and rely on techniques such as noise channel analysis, frame consistency detection, data augmentation, etc.
These detectors are all mitigation strategies after the high-quality fake content is generated and therefore are simply passive measures against \df{} attacks.
Besides, recent research shows that these detectors are vulnerable~\cite{carlini2020evading,neekhara2020adversarial}. 
To the best of our knowledge, there are no reliable defense methods against \df{} attacks yet.

This work aims to take an offensive measure to impede the generation of high-quality fake images or videos.
Poisoning attack~\cite{biggio2012poisoning} could be a solution, which compromises ML models in training phase.
If the attackers can only access elaborate poisoned faces images and use them to train their \df{} models, the synthesized faces generated by these models may not have a satisfying quality.
Rather than directly implementing poisoning attacks in high complexity, we propose an effective defense method via transformation-aware adversarial faces.
Here, we focus on GAN-based \df{} attacks as GANs are the mainstream generative models which are adopted in many state-of-the-art image translation works~\cite{isola2017image,zhu2017unpaired}. 
Specifically, we generate adversarially perturbed faces of person A based on the discriminator from a pre-trained \df{} model.
The performance of new \df{} models trained based on these adversarial faces is degraded.
This is reflected by the low quality of the synthesized faces, which  are more obviously fake to human observers and can be easily detected based on various metrics.
Therefore, the faces of person A are protected against \df{} attacks.

Our major contributions are summarized as follows:

\begin{itemize}
\item To the authors' best knowledge, it is the first offensive method that \textit{utilizes adversarial faces to defend against GAN-based \df{} attacks}.
\item We propose an adversarial face generation method to protect individuals' faces by considering random differentiable image transformations during the training of \df{} models. 
This method can consistently yield more artifacts in synthesized faces, making the recognition of the induced faked images and videos much easier.
\item We identify the increments of adversarial and edge losses as the major causes of incurring significant degradations in the quality of synthesized faces.
\item We demonstrate the effectiveness and robustness of our defense method through extensive experiments using multiple pairs of faces with different resolutions, under white-, gray-, and black-box settings based on various metrics.
\end{itemize}

\section{Related works}
\label{sec:related}

\subsection{Detection and Defense against \df{}s}

%\textbf{\df{} attacks.} 
%The very first \df{} model was built with an autoencoder based on the assumption that if warped faces can be reconstructed as the original face, then other faces can also be reconstructed similarly~\cite{deepfake_original}. So \df{} can also be called as face-swap. Although face-swap has been explored for a while~\cite{korshunova2017fast}, the traditional approaches are relatively simple and cannot adapt to the expression of the source face. \df{} emerges recently by leveraging the advances in developing generative models, such as autoencoders~\cite{kramer1991nonlinear} and GANs~\cite{goodfellow2014generative}.State-of-the-art \df{} models can grasp the features of the source face and generate fine details similar to the target face~\cite{deepfake_gan}.

\textbf{Detection of \df{} videos.}
Much of the research surrounding \df{} seeks to detect synthesized videos based on various features. 
For example, recurrent neural networks (RNNs) can be used to extract temporal information, i.e., frame-level features, for \df{} detection~\cite{guera2018deepfake}.
Yang et al.~\cite{yang2019exposing} propose to use 3D head pose estimation as a feature and adopt the support vector machine (SVM) to classify if the face is fake or not.
The use of noise analysis has also been investigated~\cite{zhou2017two}, where a two-stream NN is employed to extract both macro face features and local noise features. 
Li et al.~\cite{li2020xray} propose to use image discrepancies across the blending boundary for face forgery detection by leveraging noise analysis.
Another mainstream series of detection algorithms rely on data augmentation.
Leveraging the imperfection of synthesized videos, e.g., warping artifacts, one can effectively distinguish \df{} videos from benign counterparts~\cite{li2018exposing}.
The training of such detectors requires elaborate augmented data with similar imperfections.
The utilization of steganalysis features for data augmentation has also been explored~\cite{rossler2019faceforensics++}.

Additionally, we can use more generic image manipulation detection algorithms against \df{} videos.
A recent study indicates that convolutional neural network (CNN)-generated images are easy to spot~\cite{wang2019cnn}.
The results show that such images share common systematic flaws, which can be grasped by dedicated ML models with the assist of data augmentation.
Pixel-level image forgery detection can also be achieved.
Researchers formulate this problem as a local anomaly detection problem and solve it with the aid of an elaborate score and a long short-term memory solution~\cite{wu2019mantra}.

\textbf{Defense against \df{} videos}.
Adversarial examples against face detectors can be generated so that no valid faces can be detected, thus defending against \df{} attacks~\cite{li2019hiding}.
However, the attackers can still rely on manually extracted face regions to train \df{} models.
Therefore, such defense is essentially infeasible to provide enough protection. 
The lack of valid defenses against \df{}s motivates us to investigate this subject.

\subsection{GAN-based \df{} model}
\label{sec:model}

\begin{figure}[b]
  \centering
  \includegraphics[width=\textwidth]{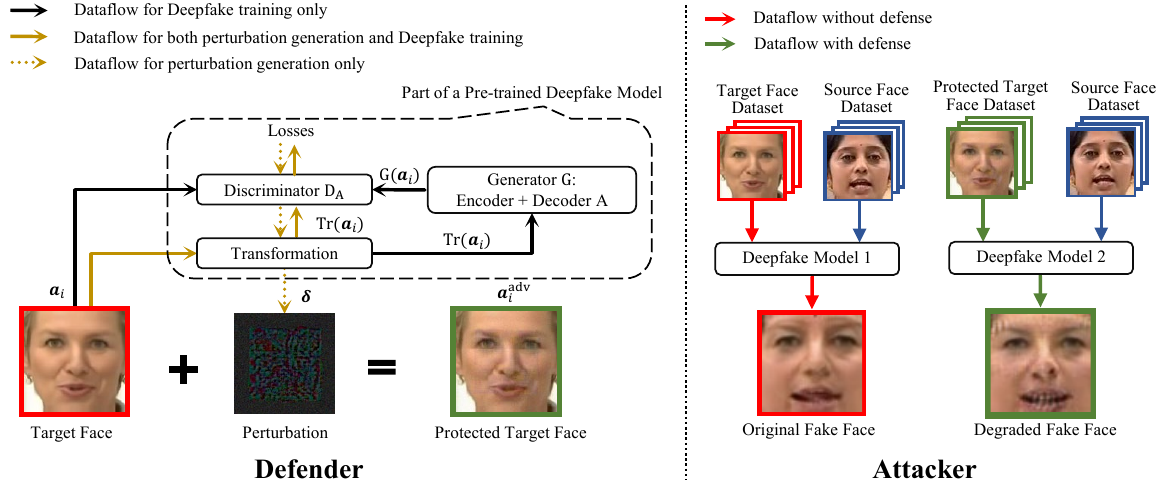} 
  \caption{An illustration of transformation-aware adversarial faces as a defense against \df{} attacks by degrading the quality of fake images or video frames. The pixel values of the perturbation in this figure is multiplied by 5 for better visual illustration. Some arrows and part of the \df{} model is omitted for clarity. Refer Equations~(\ref{equ:gan}-\ref{equ:adv_tr}) for details.}
  \label{fig:overview}
\end{figure}

The \df{} model we considered in this paper is an open-source project~\cite{deepfake_gan} based on GAN~\cite{goodfellow2014generative}.
This model is revised from the original \df{} model~\cite{deepfake_original} (which is primarily based on autoencoder~\cite{kramer1991nonlinear}) and includes many dedicated features and blocks.
The backbone of the generator is borrowed from the CycleGAN~\cite{zhu2017unpaired}.
There are also self-attention blocks in both generator and discriminator.
The detailed structure can be found in \textbf{Appendix~\ref{ap:df}}.

Given real faces $\{\bm{a}_i\}_{i=1}^{N}\in\mathcal{A}$ and $\{\bm{b}_j\}_{j=1}^{M}\in\mathcal{B}$ from two persons' face domains, the objective of training the \df{} model is to learn a mapping function $\mathrm{G}: \mathcal{B}\xrightarrow{}\mathcal{A}$, i.e., $\mathrm{G}(\bm{b}_j)\in\mathcal{A}$, with $\mathrm{F}(\bm{b}_j)\approx \mathrm{F}(\bm{a}_i)$, where $\mathrm{F}(\bm{b}_j)$ is the facial expressions of $b_j$.
Here, $\mathcal{A}$ is the target face domain (to be protected) and $\mathcal{B}$ is the source face domain.
%The mapping function $\mathrm{G}$ is usually reversible so that $\mathrm{G}: \mathcal{A}\xrightarrow{}\mathcal{B}$ is also feasible.
The dataflow of training the GAN-based \df{} model is illustrated inside the defender side of Figure~\ref{fig:overview}.
A real face $\bm{a}_i$ is transformed to $\mathrm{Tr}(\bm{a}_i)$ using a pre-determined image transformation function that includes resizing, cropping, and warping.
$\mathrm{Tr}(\bm{a}_i)$ is sent to the generator composed of a shared encoder and a dedicated decoder for $\mathcal{A}$.
The generated face $\mathrm{G}(\bm{a}_i)$ together with the real face $\bm{a}_i$ and transformed face $\mathrm{Tr}(\bm{a}_i)$ are sent to the dedicated discriminator $\mathrm{D_A}$ to calculate its loss.
%The training for $\{\bm{b}_j\}_{j=1}^{M}\in\mathcal{B}$ follows a similar way, by going through the shared encoder and its dedicated decoder and discriminator $\mathrm{D_B}$.

Various losses are adopted to improve the performance of this \df{} model as follows.
\textbf{Adversarial loss} refers to the optimization problem in Equation~(\ref{equ:gan})~\cite{goodfellow2014generative}.
This is an essential loss for GAN's training and is also the key indicator of incurred degradation in the quality of synthesized faces, which we will discuss later in Section~\ref{sec:loss_curve}.
Here, the discriminator $\mathrm{D_A}$ is trained to maximize the probability of assigning the correct label to both real samples and synthesized samples from the generator $\mathrm{G}$. 
Meanwhile, $\mathrm{G}$ is trained to minimize $\log(1-\mathrm{D_A}(\mathrm{G}(\bm{a})))$.
\begin{equation}
\label{equ:gan}
\underset{\mathrm{G}}{\min}\,\underset{\mathrm{D_A}}{\max}\,V(\mathrm{D_A},\mathrm{G})=\mathbb{E}_{\bm{a}\sim\mathcal{A}}[\log\mathrm{D_A}(\bm{a})]+\mathbb{E}_{\bm{a}\sim \mathcal{A}}[\log(1-\mathrm{D_A}(\mathrm{G}(\bm{a})))].
\end{equation}
\textbf{Reconstruction loss} is used to enhance the training of the generator to generate high-quality faces.
\textbf{Edge loss} is used to improve the details of the edges by punishing mismatches between real and fake faces.
\textbf{Cyclic loss} is evolved from the cycle consistency loss from CycleGAN~\cite{zhu2017unpaired}, which helps to regularize the structured data.
% \textbf{Perceptual loss} is first proposed in the area of neural style transfer~\cite{johnson2016perceptual}.
% In this paper, we use VGGFace~\cite{vggface} as the backbone network and extract features from four different layers to calculate this loss.
\textbf{Perceptual loss} extracts features from multiple layers of a holdout CNN model to determine the high-level differences in content or style~\cite{johnson2016perceptual}.
\section{Adversarial faces as a defense}
\label{sec:method}

\subsection{Defenders' knowledge}

The difficulty of the defense against \df{} attacks largely depends on the defenders' prior knowledge about the \df{} model and data.
Here, we consider different settings from both data and model's perspectives.
The \textit{white-box} setting indicates that the defenders have full knowledge of both data (both target faces $\{\bm{a}_i\}_{i=1}^{N}$ and source faces $\{\bm{b}_j\}_{j=1}^{M}$) and the attackers' \df{} model (including the model structure, training schemes, and hyperparameters).
The \textit{black-box} setting indicates that the defenders have zero knowledge of source faces and the \df{} model.
Note that the defenders shall have the target faces at hand because those are the faces the defenders aim to protect. %, which are usually their own faces.
The \textit{gray-box} setting is between the above two settings where the defenders have access to either source faces or the \df{} model.
In this paper, we will discuss and evaluate our defense methods under the white-box setting and then extend it to the gray- and black-box settings.

\subsection{Principle and feasibility of adversarial faces as a defense}
\label{sec:motivation}

A straight-forward offensive method to defend against \df{} attacks could be the poisoning attack that compromises the training of \df{} models.
Thus the trained \df{} models can not generate meaningful results.
However, the implementation of poisoning attacks can be very challenging even for simple feedforward NNs~\cite{yang2017generative}, let alone autoencoders~\cite{kramer1991nonlinear} and GANs~\cite{goodfellow2014generative}.
Recently, there is an interesting observation that the poisoned instances can be seen as adversarial examples~\cite{shafahi2018poison}.
Based on this phenomenon, we further explore the use of adversarial faces as poisoned samples for defending against \df{} attacks.

The key design concept behind our proposed defense method is to replace the original target faces with adversarial faces $\{\bm{a}_i^{adv}\}_{i=1}^N$ generated against the corresponding discriminator of a pre-trained \df{} model.
Consider that attackers take these adversarial faces to train a new \df{} model. 
The discriminator for the target faces ($\mathrm{D_A}$) are therefore inclined to be robust to faces from domain $\mathcal{A}$.
In other words, $\log\mathrm{D_A}(\bm{a}_i^{adv})$ is expected to increase rapidly (indicated by a lower $\mathrm{D}$ loss $\mathcal{L}_{D_A}$), which breaks the min-max game in Equation~(\ref{equ:gan}).
The training of $\mathrm{G}$ will thus be disturbed (indicated by a higher $\mathrm{G}$ loss $\mathcal{L}_G$), leading to performance degradation and low quality of the synthesized faces.
The overview of the proposed defense is shown in Figure~\ref{fig:overview}.
The defenders amend target faces with elaborate adversarial perturbations as \textit{protected target faces}.
The attackers used to generate fine fake faces with both the original target faces and source faces.
With the proposed defense, the attackers train \df{} models with the protected target faces, rather than the original target faces. 
Then the generated synthesized faces might not achieve satisfying quality and can be easily identified.
%Thus, they can only produce fake faces in a low quality which will be easily recognized by human eyes or software tools.
Note that the perturbation bound for adversarial faces should be relatively small such that the changes do not impact human's perception of the images and they can not be easily identified and removed by the attackers.
%normal usage of these faces will be affected and attackers can also easily identify them.

\subsection{Transformation-aware adversarial face generation}

A typical adversarial example can be obtained using the fast gradient sign method (FGSM)~\cite{goodfellow2014explaining} as
\begin{equation}
\label{equ:adv}
\bm{x}_{\mathrm{adv}} = \bm{x} + \alpha\, \mathrm{sign}(\nabla_{\bm{x}} \mathcal{L}(\bm{\theta}, \bm{x},y)).
\end{equation}
However, adversarial faces based on na\"ive FGSM cannot achieve satisfying defense performance.
The reason is that such adversarial faces are not robust to image transformations (e.g., resizing, cropping, warping), which are commonly applied during the training of \df{} models.
The effect of the adversarial perturbations will be compromised after these image transformation operations.

\begin{algorithm}[b]
\footnotesize
	\caption{Transformation-aware adversarial face generation using PGD.}
	\label{alg:trans}
	\begin{algorithmic}[1]
		\State \textbf{Input:} Target faces $\{\bm{a}_i\}_{i=1}^{N}\in\mathcal{A}$, adversarial faces $\{\bm{a}_i^{\mathrm{adv}}|\bm{a}_i^{\mathrm{adv}}=a_i\}_{i=1}^{N}$, source face domain $\mathcal{B}$, pre-trained \df{} model $\mathrm{M}$ based on $\mathcal{A}$ and $\mathcal{B}$, loss function of the corresponding discriminator $\mathcal{L}_{D_A}$, transformation function $\mathrm{Tr}$, iteration $Iter$, step size $\alpha$, perturbation bound $\epsilon$, the label for real faces $y^{\mathrm{real}}$.
		\For{$i$ from $1$ to $N$}
    		\For{$j$ from $1$ to $Iter$}
    		\State Calculate the adversarial perturbation: $\delta=\alpha\cdot\nabla_{\bm{a}_{i,j}^{\mathrm{adv}}}\mathcal{L}_{D_A}(\bm{\theta}, \mathrm{Tr}(\bm{a}_{i,j}^{\mathrm{adv}}), y^{\mathrm{real}})$
    		\State Clip based on $\ell_{\infty}$ norm: $\bm{a}_{i,j+1}^{\mathrm{adv}}=\mathrm{clip}(\bm{a}_{i,j}^{\mathrm{adv}}+\delta,-\epsilon,\epsilon)$
    		\EndFor
		\EndFor
		\State \textbf{Output:} Adversarial faces $\{\bm{a}_i^{\mathrm{adv}}\}_{i=1}^{N}$.
	\end{algorithmic}
\end{algorithm}

Therefore, we propose to consider the transformations to improve the effectiveness and robustness of the generated adversarial faces. 
Specifically, we apply random image transformations to the faces before sending them to the pretrained \df{} model for crafting adversarial examples.
The transformation function in this work includes resizing, affine transformation, and image remapping.
The entire process is made differentiable to utilize the automatic differentiation in deep learning frameworks such as Tensorflow~\cite{tensorflow2015-whitepaper}.
The dataflow of the adversarial face generation process is shown in Figure~\ref{fig:overview}.
Now we can generate adversarial examples against $\mathrm{D_A}$ as
\begin{equation}
\label{equ:adv_tr}
\bm{a}_i^{\mathrm{adv}} = \bm{a}_i+\alpha\, \mathrm{sign}(\nabla_{\bm{a}_i} \mathcal{L}_{D_A}(\bm{\theta}, \mathrm{Tr}(\bm{a}_i),y^{\mathrm{real}})),
\end{equation}
where $\mathrm{Tr}()$ represents the random transformation function and $y^{\mathrm{real}}$ is the label for real faces.

Algorithm~\ref{alg:trans} shows the pseudocode of the proposed transformation-aware adversarial face generation algorithm using projected gradient descent (PGD)~\cite{madry2017towards}.
Among various adversarial attacks such as FGSM, PGD, momentum iterative FGSM~\cite{dong2018boosting}, CW attack~\cite{carlini2017towards}, and universal adversarial attack~\cite{moosavi2017universal}, PGD with a multi-iteration scheme becomes the mainstream algorithm because of its simplicity and effectiveness.
The transformation function is embedded inside the PGD loop so that this function varies slightly with iterations.
Such randomness improves the robustness of the generated adversarial faces.
The transformation function can also be customized based on needs, as long as being differentiable.
Meanwhile, the algorithm can be easily incorporated with other adversarial attack methods with trivial modifications.
The comparison of the defense performance of different adversarial attack methods can be found in \textbf{Appendix~\ref{ap:at}}.

\subsection{Ensemble method and more}

The transformation-aware adversarial face generation is effective under the white-box setting.
Furthermore, we extend it to the black-box setting, under which the distribution of the source faces is unknown.
We could apply the defense based on source faces from the distribution of a random person, but this may not effectively approximate the unknown distribution.
Instead, we propose to generate adversarial faces from pre-trained \df{} models based on faces sampled from multiple distributions $\{\mathcal{C}_i\}_{i=1}^{K}$ that are different from the actual source domain $\mathcal{B}$ used by the attackers.
Algorithm~\ref{alg:etrans} shows the pseudocode of the ensemble version of the proposed method.
These ensemble adversarial faces can better approximate the optimum adversarial faces generated based on the unknown domain $\mathcal{B}$, due to the fact that adversarial examples transfer between models~\cite{tramer2017ensemble}.
Therefore, such adversarial faces can be more robust under both gray- and black-box settings. 
In addition, we propose a ``random'' method that overlaps the perturbations of random directions to the target faces.
We also clip these perturbations so that they are within the same perturbation bound $\epsilon$ (similar to PGD).
This method is thus independent of both the model and data, and follows the same procedure under all settings.

\begin{algorithm}%[tb]
\footnotesize
	\caption{Ensemble transformation-aware adversarial face generation using PGD.}
	\label{alg:etrans}
	\begin{algorithmic}[1]
		\State \textbf{Input:} Target faces $\{\bm{a}_i\}_{i=1}^{N}\in\mathcal{A}$, adversarial faces $\{\bm{a}_i^{\mathrm{adv}}|\bm{a}_i^{\mathrm{adv}}=a_i\}_{i=1}^{N}$, $K$ splits of $N$ as $\{N_k|\sum_{k=1}^K N_k = N\}_{k=1}^{K}$, face domains $\{\mathcal{C}_k\}_{k=1}^{K}$, pre-trained \df{} models $\{\mathrm{M}_k|\mathrm{M}_k\,\text{is trained with}\,\mathcal{A}, \mathcal{C}_k\}_{k=1}^{K}$, loss functions of corresponding discriminators $\{\mathcal{L}_{D_A}^k\}_{k=1}^{K}$, transformation function $\mathrm{Tr}$, iteration $Iter$, step size $\alpha$, perturbation bound $\epsilon$, the label for real faces $y^{\mathrm{real}}$.
		\For{$k$ from $1$ to $K$}
		    \For{$i_k$ from $1$ to $N_k$}
        		\For{$j$ from $1$ to $Iter$}
        		\State Calculate the adversarial perturbation: $\delta=\alpha\cdot\nabla_{\bm{a}_{(i_k,k),j}^{\mathrm{adv}}}\mathcal{L}_{D_A}^k(\bm{\theta}, \mathrm{Tr}(\bm{a}_{(i_k,k),j}^{\mathrm{adv}}), y^{\mathrm{real}})$
        		\State Clip based on $\ell_{\infty}$ norm: $\bm{a}_{(i_k,k),j+1}^{\mathrm{adv}}=\mathrm{clip}(\bm{a}_{(i_k,k),j}^{\mathrm{adv}}+\delta,-\epsilon,\epsilon)$
        		\EndFor
        	\EndFor
		\EndFor
		\State \textbf{Output:} Adversarial faces $\{\bm{a}_i^{\mathrm{adv}}|\{i\}_{i=1}^{N}\leftrightarrow\{(i_k,k)|i_k\in[1,N_k], k\in[1,K]\}\}_{i=1}^{N}$.
	\end{algorithmic}
\end{algorithm}
\section{Evaluation}
\label{sec:evaluation}

\subsection{Experiment setup}
We experiment on a subset of faces in Faceforensics++~\cite{rossler2019faceforensics++} ($\sim$400 images for each face), as the training of \df{} models is time-consuming (typically 30 GPU hours for one model).
Specifically, we randomly select four faces as the target faces to protect, where we have two male faces (M1 and M2) and two female faces (F1 and F2).
% 004: M1, 015: M2, 005 F1, 018 F2
For the male faces, we swap them with four other male faces labeled as M3, M4, M5, and M6
%000, 003, 006, 009, 
in order to achieve the best swapping results to human eyes.
Similarly, we swap the female faces with four other female faces labeled as F3, F4, F5, and F6.
%007, 008, 010, 012.
For each pair of faces (16 pairs in total), for example, M1 as the target vs. M3 as the source, we (from the attackers' perspective) train 8 different models as follows:

\textbf{Original} is the original \df{} model trained with real faces of M1 and M3.
\textbf{PGD-01} and \textbf{PGD-005} are two white-box \df{} models trained with transformation-aware adversarial faces of M1 (perturbation bound $\epsilon = 0.1$ and $0.05$, respectively) and real M3.
These adversarial faces are generated from a pre-trained \df{} model (with the same architecture) based on real M1 and M3.
\textbf{Ensemble} is similar to PGD-01. 
The difference is that the adversarial M1 are generated from pre-trained \df{} models (with the same architecture) based on real M1 and M7/F1/F4 which are different from the source faces to be swapped from (e.g., M3). 
Thus, applying Ensemble is under the gray-box setting.
\textbf{Random} is a \df{} model trained with randomly perturbed faces M1 with perturbation bound $\epsilon = 0.1$ and real faces M3.
\textbf{Lite} is a \df{} model trained with real faces M1 and M3 but with a different architecture than Original. 
This model has fewer channels (e.g., half) in most layers and serves as a lite version for limited computing resource.
\textbf{Lite-Ens} is the lite model trained with faces generated from Ensemble. 
In this case, both the data and model are unknown to the defender. 
This model is used to explore the generalization ability of our defense method under the black-box setting.
\textbf{Lite-Random} is similar to Random but trained with the lite model.

We perform the experiments on two different resolutions of the \df{} model, i.e., $64\times64$ and $128\times128$, to evaluate the robustness of our defense.
In this work, we focus on the raw swapped faces instead of the faces after post-processing because 1) post-processing can vary dramatically from project to project, and 2) current state-of-the-art detectors generally handle poorly on post-processed images/videos.
We also find that training with full adversarial faces is better than a mix of adversarial and real faces, of which the comparison results are included in \textbf{Appendix~\ref{ap:faces}}.
In this section, we present only the results on full adversarial faces, i.e., all target faces are adversarial.

\subsection{Analyzing the degradations of synthesized faces}
\label{sec:loss_curve}

\begin{figure}[b]
  \centering
  \includegraphics[width=\textwidth]{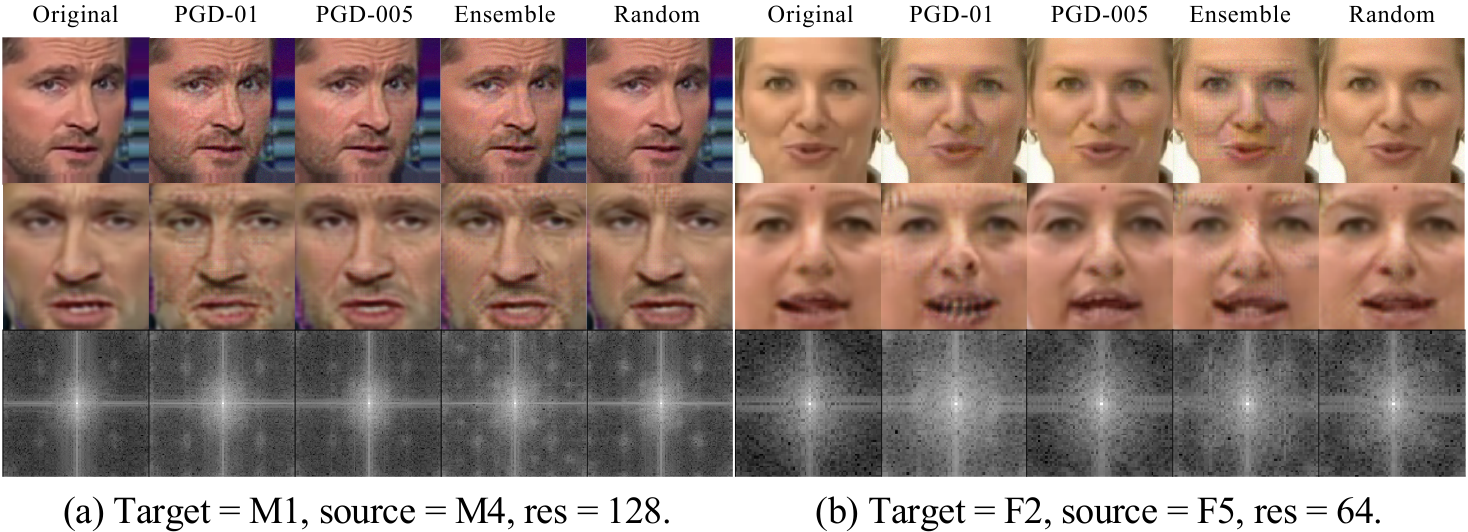}
  \caption{Face degradation comparison. For each group of results: the first row shows the target faces used for \df{} training, the second row shows the corresponding synthesized faces swapped from the source faces, and the third row shows the spectrums of the synthesized faces after FFT; columns 1-5 represent 5 models: Original, PGD-01, PGD-005, Ensemble, Random. The spectrums are shifted and scaled for better visualization, so that the zero frequency component is at the center.}
  \label{fig:face}
\end{figure}

Figure~\ref{fig:face} compares the visualization results of face degradations of two different target faces with 5 different models (see \textbf{Appendix~\ref{ap:synthesized}} for more results).
The real and protected target faces, corresponding synthesized faces, and spectrums of Fast Fourier Transform (FFT) are presented in the figure.
For both resolutions and target faces, significantly more artifacts can be observed in the synthesized faces generated by models with defense compared to the Original model without defense.
This phenomenon is also confirmed by comparing the high-frequency regions of the FFT spectrums.
The average intensities in these regions are significantly higher (lighter color) in models with defense, indicating more noise-like signals in corresponding synthesized faces.
Moreover, the spectrums sometimes show clear patterns in the high-frequency regions, especially when the resolution is 128.
Note that compared with PGD-01, PGD-005 usually incurs less degradation but also provides a lower perturbation bound, implying a trade-off between usability and security.

\begin{figure}[tb]
  \centering
  \includegraphics[width=0.95\textwidth]{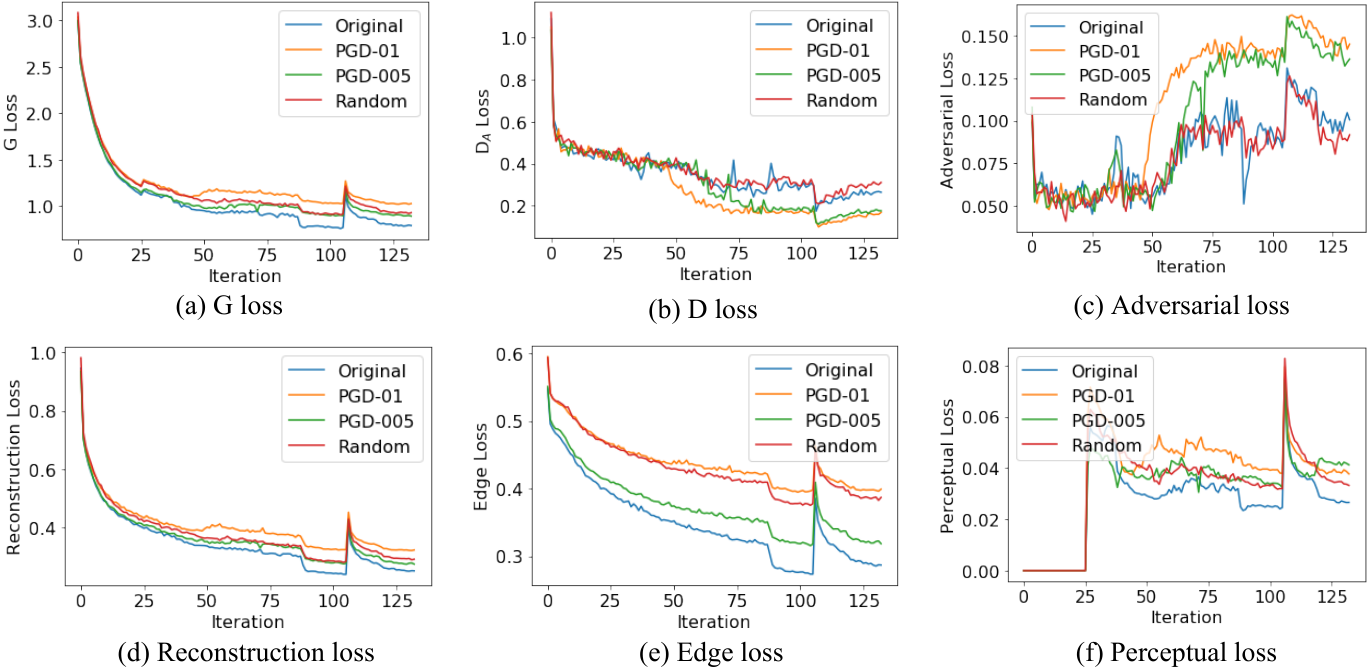} 
  \caption{Loss curves with M1/M3 as target/source under the white-box setting, resolution = 64.}
  \label{fig:loss_curve}
\end{figure}

To identify the major causes of degradation, we analyze the loss curves during the training phase of \df{} models.
Figure~\ref{fig:loss_curve} compares the losses of the four models under the white-box setting. %, including \textit{Original}, \textit{PGD-01}, \textit{PGD-005}, and \textit{Random}.
Figure~\ref{fig:loss_curve}(a-b) show that with the transformation-aware defense, the converged $\mathrm{G}$/$\mathrm{D}$ loss is higher/lower compared with Original's, which validates the explanation in Section~\ref{sec:motivation}.
% Figure~\ref{fig:face} and Figure~\ref{fig:loss_curve}(a) indicate that higher converged $\mathrm{G}$ loss usually leads to a higher degradation.
Additionally, a model with higher converged $\mathrm{G}$ loss in Figure~\ref{fig:loss_curve}(a) usually results in more degradation in the quality of synthesized faces.
For instance, the degradation of PGD-01 synthesized faces is more visible than that of Random as shown in Figure~\ref{fig:face} for both target faces.
%Figure~\ref{fig:face} (e.g., PGD-01 has higher $\mathrm{G}$ loss than Random, and the degradation of PGD-01 synthesized faces is more visible than that of Random as shown in Figure~\ref{fig:face} for both target faces).
Figure~\ref{fig:loss_curve}(c-d) present the breakdowns of the $\mathrm{G}$ loss (cyclic loss is not included because it is rather trivial) as mentioned in Section~\ref{sec:model}.
It can be easily observed that adversarial loss and edge loss are two strong indicators of degradation.
PGD-01 achieves the highest converged losses in both cases and leads to more degradation in general.
PGD-005 has similar adversarial loss but lower edge loss and incurs less degradation than PGD-01.
Random achieves similar degradation to PGD-005, however, the adversarial loss is even slightly lower than Original, indicating a different behavior from PGDs.
The edge loss of Random, on the other hand, is only slightly lower than PGD-01, possibly due to the same perturbation bound.
For reconstruction and perceptual losses, the differences are less significant.

\begin{table}[tb]
\footnotesize
  \caption{Average intensity of high-frequency regions (AIH) of FFT spectrums.}
  \label{tab:fft}
  \centering
  \begin{tabular}{llllllllll}
    \toprule
    \multirow{2}{*}{Setting} & Target face & \multicolumn{2}{c}{M1} & \multicolumn{2}{c}{F1} & \multicolumn{2}{c}{M2} & \multicolumn{2}{c}{F2}\\
    \cmidrule(lr){2-10}
    & Resolution &64 & 128 & 64 & 128 &64 & 128 & 64 & 128\\
    \midrule
    \multirow{4}{*}{White-box} & Original & 65.34 & 72.15 & 50.54 & 54.39 & 62.13 & 71.41 & 49.41 & 53.14\\
    & PGD-01   & \textbf{89.49} & \textbf{96.66} & \textbf{90.82} & \textbf{105.54} & \textbf{76.22} & 86.82 & \textbf{79.54} & \textbf{116.65} \\
    & PGD-005  & 68.36 & 84.34 & 65.13 & 60.44 & 69.56 & 81.54 & 57.24 & 64.49 \\
    & Random    & 70.13 & 95.21 & 58.61 & 83.98 & 74.75 & \textbf{96.01} & 65.63 & 70.46 \\
    \midrule
    Gray-box & Ensemble & 86.72 & 105.93 & 91.41 & 88.15 & 81.37 & 87.61 & 90.17 & 85.43 \\
    \midrule
    \multirow{3}{*}{Black-box} & Lite & 42.31 & 54.37 & 37.57 & 52.64 & \textbf{43.04} & 54.00 & 33.46 & 53.82\\
    & Lite-Random & 41.82 & 56.37 & 35.68 & 55.02 & 39.83 & \textbf{58.65} & 34.07 & \textbf{57.95} \\
    & Lite-Ens & \textbf{46.31} & \textbf{56.39} & \textbf{40.63} & \textbf{55.36} & 38.74 & 53.86 & \textbf{35.94} & 53.25 \\
    \bottomrule
  \end{tabular}
\end{table}

We quantitatively compare the degradation of synthesized faces with the average intensity of high-frequency regions (AIH) of FFT spectrums: $\mathrm{AIH} = \frac{1}{(r-40)\cdot(c-40)}\sum_{i=20}^{r-19}\sum_{j=20}^{c-19}s_{i,j}$,
% \begin{equation}
%     \mathrm{AIH} = \frac{1}{(r-40)\cdot(c-40)}\sum_{i=20}^{r-19}\sum_{j=20}^{c-19}s_{i,j},
% \end{equation}
where $r$/$c$ is the number of rows/columns in the spectrum, and $s_{i,j}$ is the intensity of one element in the spectrum. Here, we choose to average over the $(r-40)\times(c-40)$ center high-frequency region. 
The spectrum is not shifted for the simplicity of calculation.
The results are summarized in Table~\ref{tab:fft}, in which a higher AIH indicates more degradation.
For each target face (e.g., M1), we calculate the average AIH of swapping with 4 source faces (e.g., M3, M4, M5, M6), in both resolutions.
%As shown in \ref{tab:mantranet}, 
PGD-01 and Ensemble achieve the highest AIH (7 out of 8 cases) in general, indicating the most significant degradation. PGD-005 and Random are less effective, and Original always has the lowest AIH.
On average, PGD-01, PGD-005, Random, and Ensemble increase the AIH by $59\%$, $16\%$, $24\%$, and $52\%$, compared with Original.
For lite models, Lite-Ens is the most effective, with the highest AIH in 5 out of 8 cases.
This means that the ensemble method is indeed effective under the black-box setting.

\subsection{Evaluation using generic image manipulation detector}
\begin{figure}[tb]
  \centering
  \includegraphics[width=\textwidth]{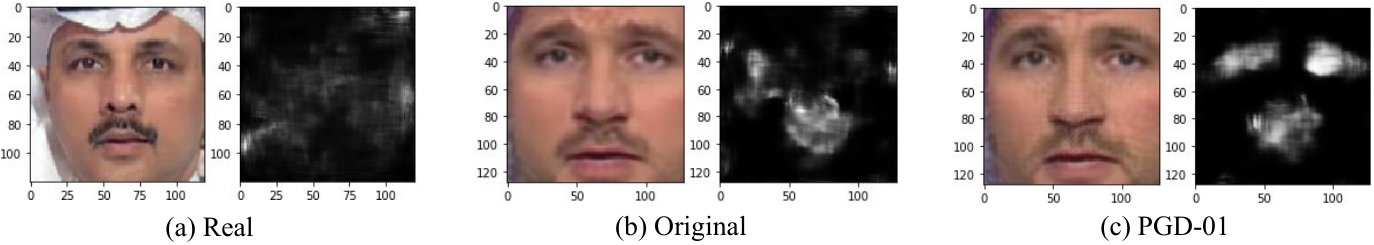} 
  \caption{ManTra-Net's detection masks of different faces.}
  \label{fig:mantranet}
\end{figure}

We also evaluate our defense using a state-of-the-art generic image forgery detection model -- ManTra-Net~\cite{wu2019mantra}. %, to evaluate how the defense will affect such detectors.
ManTra-Net generates a pixel-level detection mask reflecting the probability of manipulation in the original image.
Figure~\ref{fig:mantranet} shows the ManTra-Net masks of different synthesized faces.
More results can be found in \textbf{Appendix~\ref{ap:mantranet}}.
%As shown in Figure~\ref{fig:mantranet}(a), 
For real faces cropped from video frames, ManTra-Net tends to generate noise-like patterns, indicating no specific focus.
For synthesized faces generated by Original model %(e.g., Figure~\ref{fig:mantranet}(b)), 
ManTra-Net starts to focus on the center region of the face including the eyes and mouth.
Interestingly, for synthesized faces generated by PGD-01 model, %(e.g.,Figure~\ref{fig:mantranet}(c)), 
the detector is able to clearly segment out the region of eyes and mouth in the corresponding masks.

However, there lacks a standard way to quantify the result and determine if an image is manipulated or not.
Based on our observation of ManTra-Net masks, we propose to use the average top intensities (ATI) to filter out the background noise for classification purpose: $\mathrm{ATI} = \frac{1}{r\cdot c}\sum_{i=1}^{0.02\cdot r\cdot c}m_i$,
% \begin{equation}
%     \mathrm{ATI} = \frac{1}{r\cdot c}\sum_{i=1}^{0.02\cdot r\cdot c}m_i,
% \end{equation}
where $r$/$c$ is the number of rows/columns in the mask, $m_i\in M=\mathrm{sort}(\mathrm{flatten}(mask))$, and we choose to average the top $2\%$ intensities.
Table~\ref{tab:mantranet} summarizes the ATI results.
For each target face, we show the average ATI over 4 source faces.
%Similar to Table~\ref{tab:fft}, 
Under the white-box setting, PGD-01 dramatically increases the scores (by $28\%$ on average) compared to Original.
Random does not bring this much of increment because it has the less converged $\mathrm{G}$ loss as discussed in Section~\ref{sec:loss_curve}.
%Under the black-box setting, even though the results are relatively mixed, Lite-Ens still slightly improves the ATIs than Lite.
The results are relatively mixed under the black-box setting, but Lite-Ens still slightly improves the ATIs than Lite.
In summary, we show that synthesized faces with our defense method are easier to be detected using ManTra-Net.

\begin{table}[tb]
\footnotesize
  \caption{Average top intensities (ATI) of ManTra-Net masks.}
  \label{tab:mantranet}
  \centering
  \begin{tabular}{llllllllll}
    \toprule
     \multirow{2}{*}{Setting} & Target face & \multicolumn{2}{c}{M1} & \multicolumn{2}{c}{F1} & \multicolumn{2}{c}{M2} & \multicolumn{2}{c}{F2}\\
    \cmidrule(lr){2-10}
    &Resolution & 64 & 128 & 64 & 128 &64 & 128 & 64 & 128\\
    \midrule
    \multirow{4}{*}{White-box} & Original & 0.482 & 0.641 & 0.592 & 0.874 & 0.575 & 0.537 & 0.543 & 0.570 \\
    & PGD-01   & \textbf{0.724} & 0.790 & 0.446 & \textbf{0.880} & \textbf{0.811} & \textbf{0.894} & 0.612 & \textbf{0.900} \\
    & PGD-005  & 0.596 & 0.809 & \textbf{0.659} & 0.702 & 0.649 & 0.641 & \textbf{0.679} & 0.653 \\
    & Random    & 0.572 & \textbf{0.813} & 0.649 & 0.818 & 0.405 & 0.887 & 0.642 & 0.619 \\
    \midrule
    Gray-box & Ensemble & 0.582 & 0.803 & 0.670 & 0.778 & 0.881 & 0.784 & 0.766 & 0.752 \\
    \midrule
    \multirow{3}{*}{Black-box} & Lite & 0.281 & 0.373 & 0.255 & \textbf{0.335} & \textbf{0.354} & \textbf{0.436} & 0.251 & 0.289\\
    & Lite-Random & 0.309 & 0.333 & 0.312 & 0.331 & 0.272 & 0.347 & \textbf{0.257} & 0.270 \\
    & Lite-Ens & \textbf{0.337} & \textbf{0.374} & 0.293 & 0.329 & 0.320 & 0.377 & 0.249 & \textbf{0.316} \\
    \bottomrule
  \end{tabular}
\end{table}

\section{Conclusion}
\label{sec:conclusion}
In this paper, we propose to defend against \df{} attacks via transformation-aware adversarial faces.
We show that training a \df{} model with adversarially perturbed face images can lead to a significant degradation in the quality of synthesized faces.
This degradation can be visually detectable and easily identified by various metrics. We also identify the adversarial and edge losses as the major indicators of such degradation. 
Extensive experiment results of multiple faces under white-box, gray-box, and black-box settings demonstrate the effectiveness and robustness of our defense method based on various metrics.

\section*{Broader Impact}
% {\RED Authors are required to include a statement of the broader impact of their work, including its ethical aspects and future societal consequences. 
% Authors should discuss both positive and negative outcomes, if any. For instance, authors should discuss a) 
% who may benefit from this research, b) who may be put at disadvantage from this research, c) what are the consequences of failure of the system, and d) whether the task/method leverages
% biases in the data. If authors believe this is not applicable to them, authors can simply state this.}

\df{} can have potentially drastic security consequences if applied inappropriately~\cite{Chesney2019}. 
It pushes the need for identity protection to the next level.
The proposed methods can help defend against such security threats by degrading the performance of the GAN-based \df{} models.
Celebrities or even normal individuals can benefit from this defense.
We also see opportunities for research applying our methods to beneficial purposes, such as investigating whether adversarial examples could protect audio data from \df{} manipulation. 
% The potential risks of applying the proposed methods include: (1) if the \df{} model is retrained over time with more unprotected facial images,  or new types of \df{} models are developed, this could lead to a false sense of security; and (2) applications that learn from facial images run the risk of performance decrease.
However, there are also potential risks of applying the proposed methods.
For example, if the \df{} model is retrained over time with more unprotected facial images, or new types of \df{} models are developed, the attackers may still generate high-quality fake faces. 
This could lead to a false sense of security and unfavorable impacts on the defenders.
Additionally, applications that learn from facial images run the risk of performance decrease.
On the other hand, many industries can benefit from \df{}, such as film industry, entertainment and games, educational media, and so on~\cite{deepfake_pos}. Such benign uses of \df{} may be affected by adversarial faces.
%Another potential negative outcome is that bad actors can leverage this defense against legitimate use of \df{} such as entertainment or movie editing~\cite{deepfake_movie}.

\bibliographystyle{abbrv}
\bibliography{ref}

\newpage
\appendix

\section{Architectures of the GAN-based \df{} model}
\label{ap:df}
The generative network used in this work was revised from CycleGAN~\cite{zhu2017unpaired}.
Here are notations used in defining architectures.
``c3s1-k'' denotes a $3\times3$ Convolution-InstanceNorm-ReLU layer with k filters and stride 1.
``sa'' is the self-attention layer in SAGAN~\cite{zhang2018self}.
``up3-k'' denotes an upscale block that consists of a $3\times3$ convolutional layer with k filters and a PixelShuffle layer.
``Rk'' denotes a residual block that consists of two $3\times3$ convolutional layers, both of which have k filters.
``Dk'' denotes a dense layer with the output dimension as k.
Below are the generator and discriminator architectures.

\textbf{Generator architecture}:\\
Encoder: c3s1-64, c3s2-128, c3s2-256, sa, c3s2-512, sa, c3s2-1024, D1024, D16384, up4-512.\\
Decoder: up4-256, up4-128, sa, up4-64, R64, sa, c5s1-3.

\textbf{Discriminator architecture}:\\
c3s2-64, c3s2-128, sa, c3s2-256, sa, c5s1-1.

\section{Comparison among various adversarial attacks: PGD, FGSM, MITER, CW, and Universal }
\label{ap:at}

\begin{figure}[h]
  \centering
  \includegraphics[width=\textwidth]{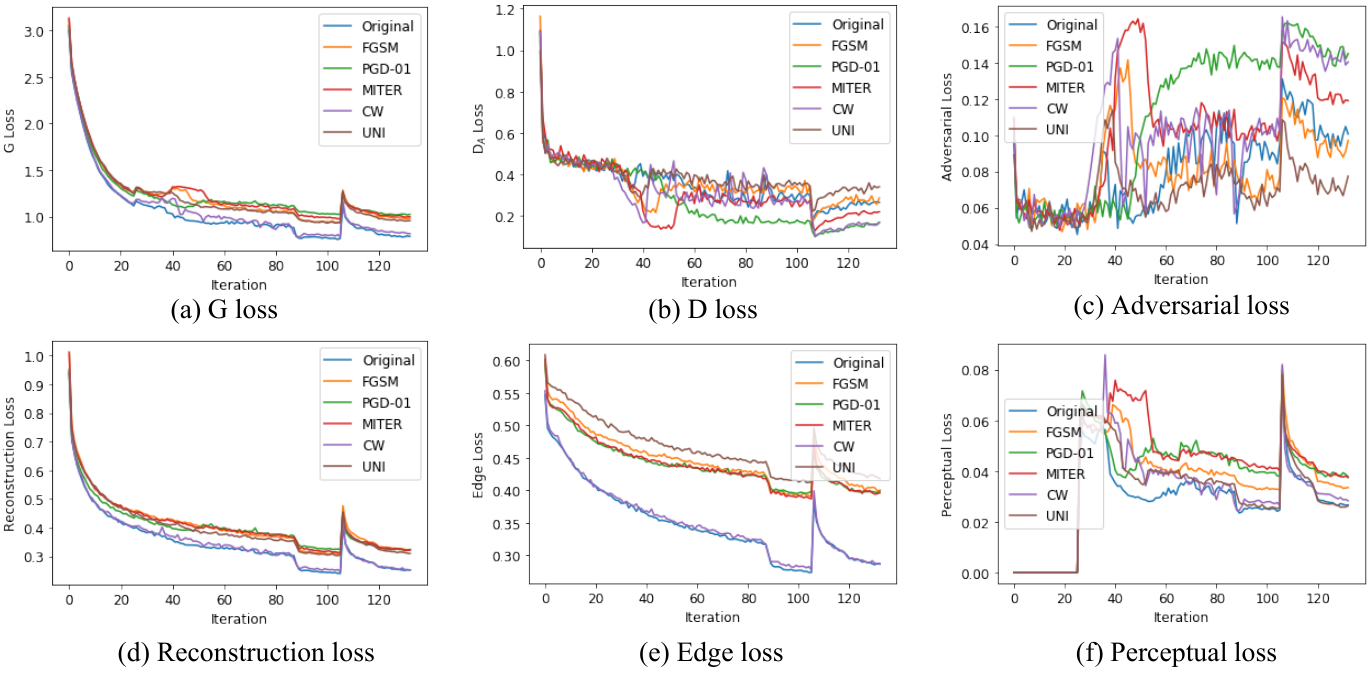} 
  \caption{Loss curves of Original, FGSM~\cite{goodfellow2014explaining}, PGD-01~\cite{madry2017towards}, momentum iterative FGSM (MITER)~\cite{dong2018boosting}, CW attack (CW)~\cite{carlini2017towards}, and universal adversarial attack (UNI)~\cite{moosavi2017universal}, with M1/M3 as target/source under the white-box setting, resolution = 64.}
  \label{fig:loss_curve_attacks}
\end{figure}

In addition to the PGD-01 attack discussed in the main content, we evaluate more adversarial attacks.
Figure~\ref{fig:loss_curve_attacks} summarizes the comparison results. 
%We compare more adversarial attacks besides the adopted PGD-01 in Figure~\ref{fig:loss_curve_attacks}.
Here, we present the loss curve comparison in a similar way to that in Section~\ref{sec:loss_curve}.
According to the previous discussion, we focus mainly on the adversarial loss and the edge loss during the comparison.
As can be seen that PGD-01 and CW achieve the highest adversarial loss, while CW has almost the same (lowest) edge loss as Original.
The highest edge loss is obtained by UNI, which also has the lowest adversarial loss. 
PGD-01, MITER, and FGSM achieve similar edge losses which are slightly lower than UNI.
MITER and FGSM are in the middle for both losses.
Overall, the results show that PGD-01 achieves the highest $\mathrm{G}$ loss and also performs well in other evaluations.

\newpage

\section{Comparison of different percentages of adversarial faces}
\label{ap:faces}

\begin{figure}[h]
  \centering
  \includegraphics[width=\textwidth]{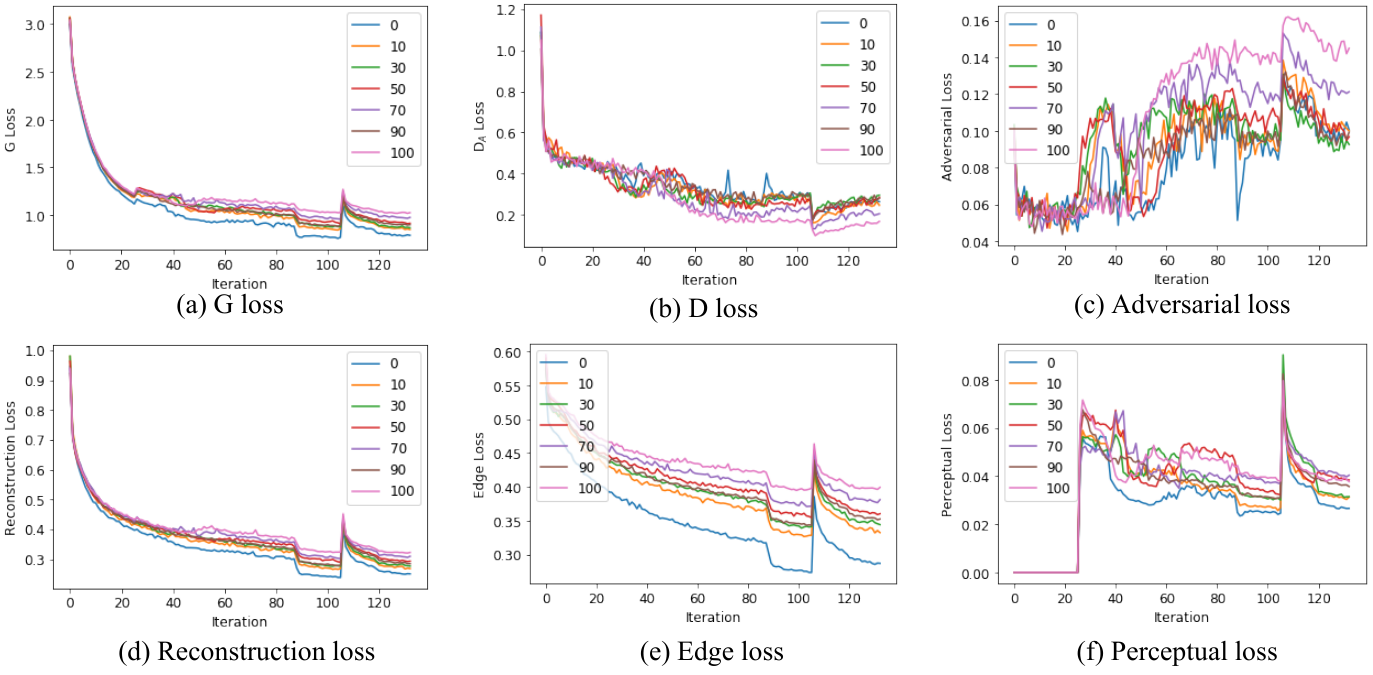} 
  \caption{Loss curves with different percentages of adversarial faces (e.g., 10 means $10\%$ of the target faces are adversarial faces), with M1/M3 as target/source under the white-box setting, resolution = 64.}
  \label{fig:loss_curve_mix}
\end{figure}

We compare the results of training \df{} models with different percentages of adversarial faces in Figure~\ref{fig:loss_curve_mix}.
For example, $30\%$ means a \df{} model trained with target faces consisting of $30\%$ adversarial faces and $70\%$ benign faces.
Generally, a higher percentage of adversarial faces incurs a higher $\mathrm{G}$ loss.
In this specific case, $90\%$ achieves lower $\mathrm{G}$ loss which may due to the randomness in the selection of adversarial faces.
As a result, we choose to use $100\%$ adversarial faces which is indeed a feasible solution because we just need to add perturbations to all target faces.
In real-world scenarios, this result also indicates that the defense will be partially compromised if the attackers have access to benign faces.

\newpage

\section{More results of synthesized faces with FFT spectrums}
\label{ap:synthesized}

\begin{figure}[h]
  \centering
  \includegraphics[width=0.9\textwidth]{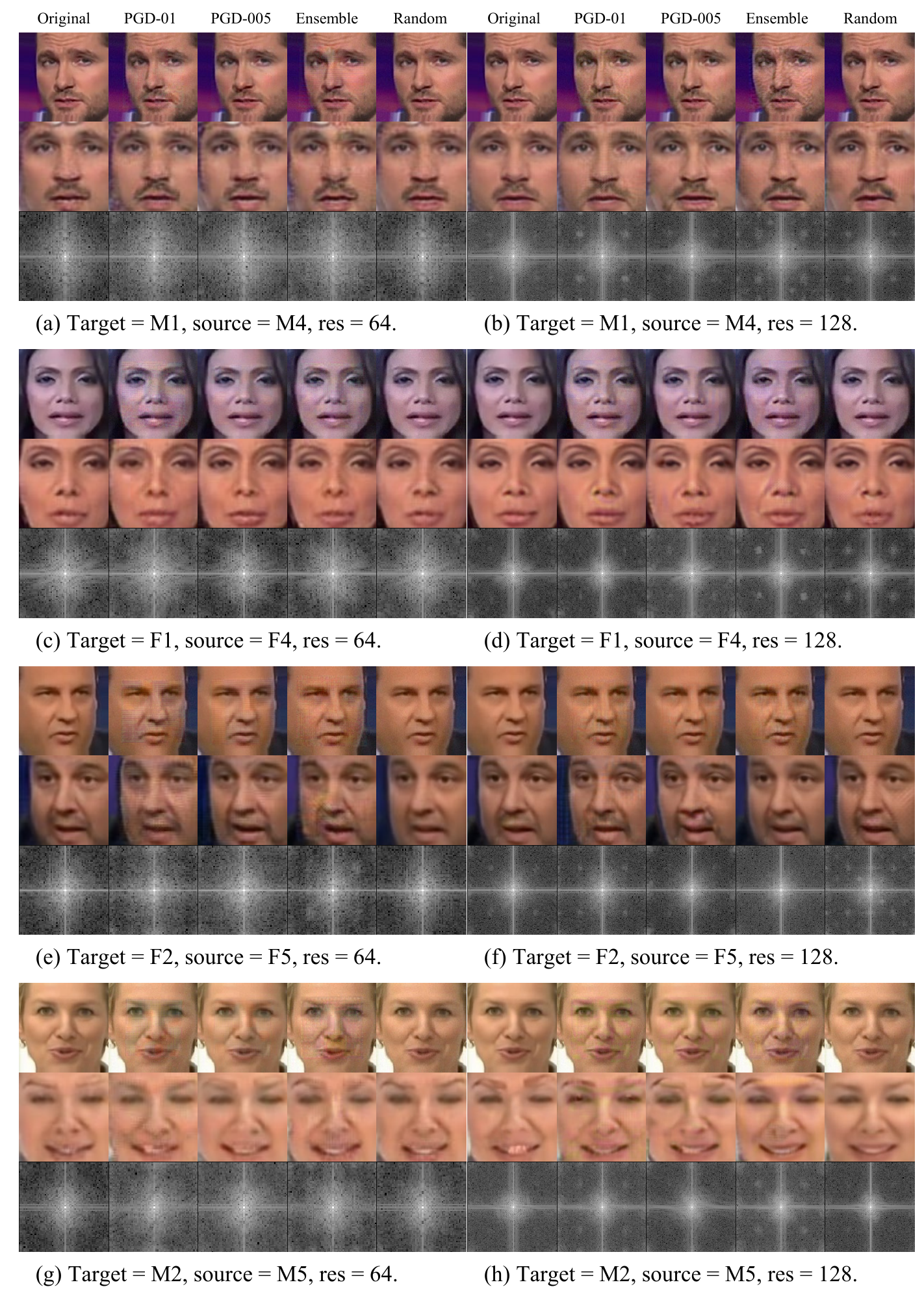}
  \caption{Face degradation comparison. For each group of results: the first row shows the target faces used for \df{} training, the second row shows the corresponding synthesized faces swapped from the source faces, and the third row shows the spectrums of the synthesized faces after FFT; columns 1-5 represent 5 models: Original, PGD-01, PGD-005, Ensemble, Random. The spectrums are shifted and scaled for better visualization, so that the zero frequency component is at the center.}
  \label{fig:face_fft_appendix}
\end{figure}

\newpage
\section{More results of ManTra-Net detection mask}
\label{ap:mantranet}

\begin{figure}[h]
  \centering
  \includegraphics[width=0.9\textwidth]{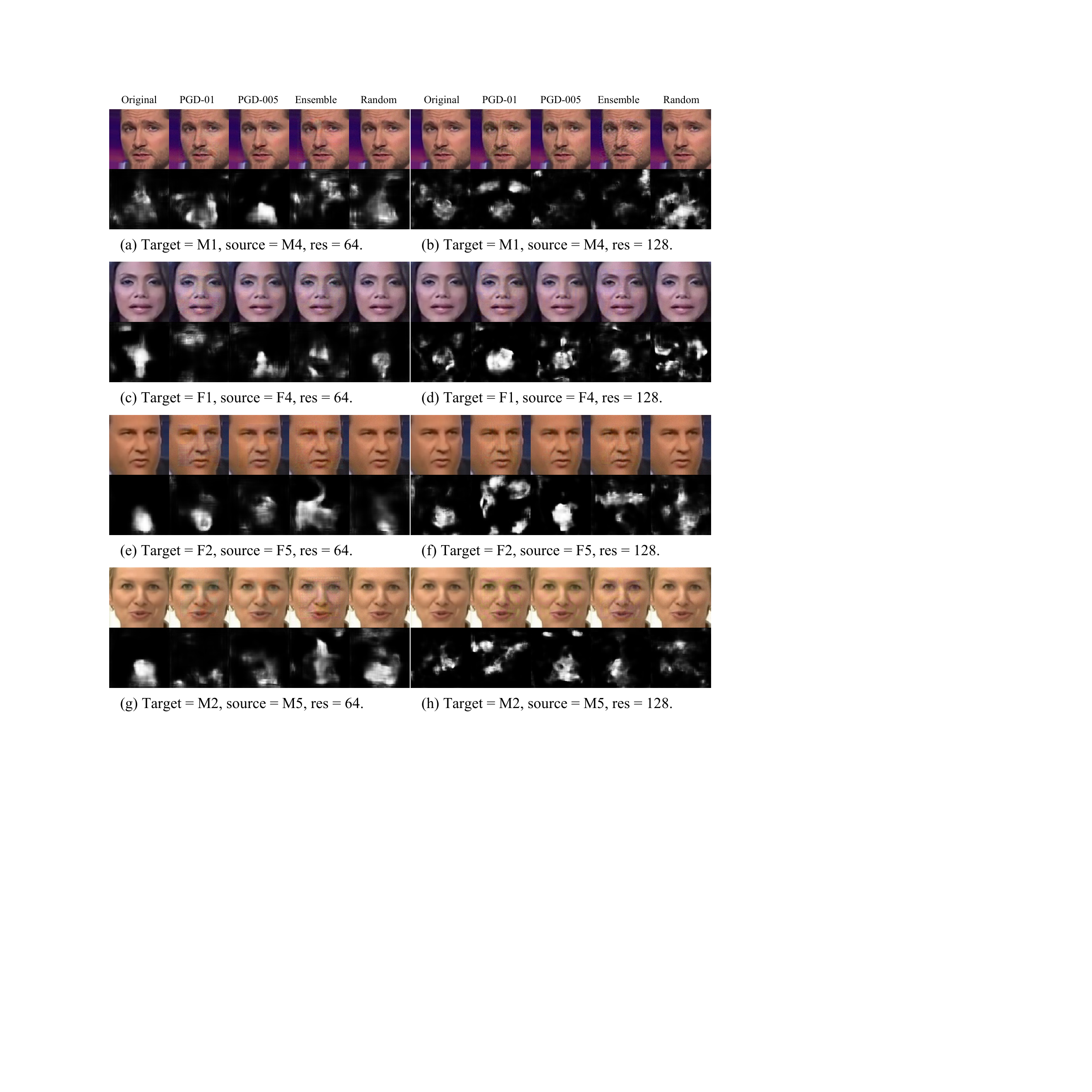}
  \caption{ManTra-Net detection mask comparison. For each group of results: the first row shows the target faces used for \df{} training, the second row shows the corresponding ManTra-Net masks; columns 1-5 represent 5 models: Original, PGD-01, PGD-005, Ensemble, Random.}
  \label{fig:face_mask_appendix}
\end{figure}

\end{document}